\pdfoutput=1
\documentclass[10pt,twocolumn,letterpaper]{article}
\usepackage[letterpaper,textwidth=178mm,textheight=229mm,centering,columnsep=6mm]{geometry}
\usepackage{times}
\usepackage{amsmath,graphicx}
\newenvironment{keywords}{\par\smallskip\noindent\textbf{\textit{Index Terms}---}\ignorespaces}{\par}
\makeatletter
\renewcommand{\section}{\@startsection{section}{1}{\z@}{8pt plus 2pt minus 2pt}{4pt plus 1pt minus 1pt}{\normalfont\normalsize\bfseries}}
\renewcommand{\subsection}{\@startsection{subsection}{2}{\z@}{7pt plus 2pt minus 2pt}{3pt plus 1pt minus 1pt}{\normalfont\normalsize\bfseries}}
\renewcommand{\@maketitle}{\newpage\null\vskip 1em\begin{center}
{\large\bfseries\@title\par}\vskip 1em
{\normalsize\begin{tabular}[t]{c}\@author\end{tabular}\par}
\end{center}\par\vskip 1em}
\makeatother
\usepackage{amsfonts}
\usepackage[T1]{fontenc}
\usepackage{booktabs}

\usepackage{enumitem}
\setlist{nosep, leftmargin=12pt}

\graphicspath{{fig/arch_diagram/}{fig/results_fig/}}

\newcommand{\ci}[2]{[#1,\,#2]}

\let\oldthebibliography\thebibliography
\renewcommand{\thebibliography}[1]{\oldthebibliography{#1}\setlength{\itemsep}{0pt}\setlength{\parsep}{0pt}}

\title{Band2Net: Frequency-to-Network Transport Distillation for
       fMRI-Privileged, EEG-Only Session-Condition Decoding}
\author{\normalsize Nishi Kanta Paul$^{\star}$ \qquad Md Shihabul Islam Shovo$^{\star}$ \qquad Dwarikanath Mahapatra$^{\dagger}$ \\
\normalsize $^{\star}$NIMISHES Lab, Dhaka, Bangladesh \qquad
$^{\dagger}$Khalifa University, Abu Dhabi, UAE \\
\small nishikantapaul108@gmail.com, shihabul900@gmail.com, dwarikanath.mahapatra@ku.ac.ae}
\date{}
\begin{document}
\renewcommand{\baselinestretch}{0.95}
\let\normalsize\small
\normalsize
\setlength{\abovedisplayskip}{4pt plus 1pt minus 1pt}
\setlength{\belowdisplayskip}{4pt plus 1pt minus 1pt}
\maketitle
\begin{abstract}
Functional MRI resolves large-scale brain networks but is expensive and
stationary, whereas EEG is portable but poorly localised. We treat fMRI as
privileged information, available during training and absent at deployment,
with EEG and fMRI acquired from the same participants under matched conditions
but not simultaneously. Band2Net encodes EEG into five frequency-specific
connectivity graphs and fMRI into seven canonical network tokens from a frozen
teacher, and solves a balanced entropic transport problem between the two token
sets that gives each EEG band its own fMRI target. In subject-grouped
cross-validation on the PsiConnect psilocybin
data set, Band2Net reaches
$0.801$ ROC-AUC for EEG-only session-condition decoding, $5.0$ AUC points
($95\%$ CI $1.8$--$8.2$) above an EEG-only student. Although its margins
over a single shared plan, a KL-relaxed plan and label-free network descriptors
remain unresolved, secondary comparisons favour band-specific transported
targets over logit, global-feature, contrastive and pooled-target distillation,
though not after Holm correction.
\end{abstract}


\section{Introduction}

fMRI reveals large-scale functional networks and underlies the canonical
parcellations of network neuroscience~\cite{yeo2011}. EEG resolves
millisecond rhythms and is inexpensive, portable and repeatable, yet has no
explicit notion of a cortical network and localises poorly in sensor space. The two connectomes are related
but far from redundant, and their coupling is present in every frequency band
with band-dependent topology~\cite{deligianni2014,wirsich2020,wirsich2021}.

This asymmetry matters when a brain state must be tracked outside a scanner.
Psilocybin enhances between-network communication and reduces default-mode
integrity~\cite{daws2022} while inducing broadband desynchronisation most
pronounced in the alpha range~\cite{muthukumaraswamy2013}; following such
network-level effects across a dosing day or repeated trial visits is feasible
with EEG and impractical with fMRI. We therefore use pre- vs.\ post-dose
session decoding, whose label the experimenter knows, as a proxy benchmark: if
fMRI network structure transfers into an EEG-only model, that model should
separate the two states better than one trained on EEG alone.

Current approaches leave three gaps. (i)~Fusion
models~\cite{yap2025brainmgf} need fMRI at test time, and
EEG-to-fMRI synthesis~\cite{yao2025catd} avoids this only by
regressing regional BOLD signals from simultaneous recordings. (ii)~Privileged
learning lets fMRI shape training and then discards
it~\cite{vapnik2009,lopezpaz2016}, but logit~\cite{hinton2015} and pooled-feature
distillation~\cite{romero2015fitnets}, including from fused fMRI--EEG
teachers in CMD-AD~\cite{lin2026cmdad}, reduce the teacher to targets with a fixed
correspondence and discard the network structure that makes fMRI worth
acquiring; FRIST~\cite{zhang2026frist} learns fMRI-informed spectral projections
for EEG-only decoding, but from simultaneous EEG--fMRI. (iii)~A band-to-network
mapping cannot be specified a priori: coupling is band dependent and may differ
between individuals~\cite{deligianni2014,wirsich2020}; we test whether
per-observation routing is needed. Optimal transport compares structured
sets; it has served distillation~\cite{chen2021wcord} and privileged multimodal
distillation between examples in PKDOT~\cite{aslam2024pkdot,aslam2024mtpkdot},
but not routing from
EEG bands to fMRI networks.

We therefore ask: given EEG and fMRI from the same participants under matched
conditions but not simultaneously, can the network-level organisation captured
by fMRI be transferred into a decoder that uses EEG alone, and must that
transfer be frequency-band specific? We solve a balanced entropic transport
problem between five EEG band representations and seven canonical fMRI network
tokens for every paired observation, and use the plan to give each band its own
privileged target. Unlike PKDOT (between examples) and CMD-AD (fixed
scale-to-scale matching), routing is learned within each observation, over
bands against networks.

Our main contributions are as follows:
\begin{itemize}
  \item We cast fMRI-privileged EEG learning from separately acquired,
  state-matched recordings as balanced frequency-to-network transport, giving
  each band its own fMRI target.
  \item We distil these targets into an EEG-only student through an asymmetric
  stop-gradient that keeps the teacher space fixed.
  \item We show under subject-grouped validation that fMRI privilege improves
  EEG-only decoding, and that label-free network descriptors recover most of
  the gain without a learned teacher.
\end{itemize}

\begin{figure*}[t]
  \centering
  \includegraphics[width=0.68\textwidth]{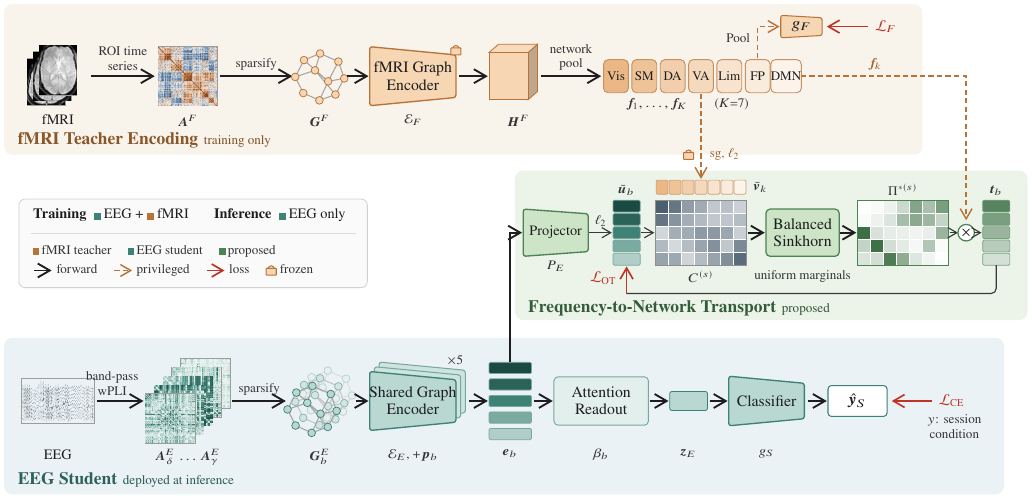}
  \caption{Band2Net. Five band-wise EEG connectivity graphs and seven canonical
  fMRI network tokens are matched by a balanced transport plan
  $\Pi^{*}$ computed per paired observation, whose transported targets
  $t_b$ supervise the EEG student through a stop-gradient; the
  frozen fMRI branch is discarded at inference.}
  \label{fig:arch}
\end{figure*}

\begin{table}[t]
\centering
\caption{EEG-only decoding of session condition on the $112$ session-level
predictions (Sec.~\ref{ssec:protocol}). AUC with 95\% subject-level bootstrap
CI. Gain: paired AUC difference, Band2Net minus method, in AUC points
($\mathrm{AUC}\times100$) with 95\% CI; $^\dagger$primary comparison;
$^{*}$significant after Holm correction over the $16$ secondary comparisons
in this table. BA:
balanced accuracy, F1: macro-F1 (\%). Per-seed AUC SD is at most
$0.012$ for every method. The fMRI teacher is not an EEG-only model.}
\label{tab:main}
\footnotesize
\setlength{\tabcolsep}{2pt}
\begin{tabular}{@{}l c c c c@{}}
\toprule
Method & ROC-AUC & Gain & BA & F1 \\
\midrule
\multicolumn{5}{@{}l}{\itshape EEG-only training} \\
Band power + LR & 0.694\,\ci{0.640}{0.746} & 10.7\,\ci{6.3}{15.2}$^{*}$ & 64.3 & 63.9 \\
EEG wPLI FC + SVM & 0.728\,\ci{0.676}{0.778} & 7.3\,\ci{3.2}{11.5}$^{*}$ & 66.1 & 65.7 \\
Riemannian TS + LR & 0.739\,\ci{0.688}{0.788} & 6.2\,\ci{2.0}{10.5} & 67.0 & 66.6 \\
EEG-only GNN & 0.751\,\ci{0.701}{0.799} & 5.0\,\ci{1.8}{8.2}$^\dagger$ & 68.8 & 68.3 \\
\addlinespace[1pt]
\multicolumn{5}{@{}l}{\itshape fMRI-privileged training, EEG-only inference} \\
Logit KD & 0.759\,\ci{0.709}{0.806} & 4.2\,\ci{1.3}{7.1} & 69.6 & 69.2 \\
Global feature KD & 0.768\,\ci{0.719}{0.814} & 3.3\,\ci{0.6}{6.0} & 70.5 & 70.1 \\
CRD & 0.771\,\ci{0.722}{0.817} & 3.0\,\ci{0.2}{5.8} & 70.5 & 70.1 \\
Pooled-target & 0.772\,\ci{0.723}{0.818} & 2.9\,\ci{0.1}{5.7} & 70.5 & 70.2 \\
\addlinespace[1pt]
\multicolumn{5}{@{}l}{\itshape Band2Net ablations} \\
Fixed random plan & 0.774\,\ci{0.725}{0.820} & 2.7\,\ci{0.4}{5.1} & 70.5 & 70.2 \\
Row-softmax plan & 0.776\,\ci{0.727}{0.822} & 2.5\,\ci{0.3}{4.7} & 71.4 & 71.0 \\
Shared plan & 0.787\,\ci{0.739}{0.832} & 1.4\,\ci{$-$0.7}{3.5} & 72.3 & 71.9 \\
Unbalanced plan & 0.791\,\ci{0.744}{0.836} & 1.0\,\ci{$-$0.9}{2.9} & 71.4 & 71.1 \\
Descriptor targets & 0.793\,\ci{0.746}{0.837} & 0.8\,\ci{$-$1.4}{3.1} & 72.3 & 72.1 \\
Cross-fitted tokens & 0.794\,\ci{0.747}{0.839} & 0.7\,\ci{$-$1.1}{2.5} & 72.3 & 71.9 \\
\textbf{Band2Net (proposed)} & \textbf{0.801\,\ci{0.755}{0.844}} & -- & \textbf{73.2} & \textbf{72.8} \\
\addlinespace[1pt]
\multicolumn{5}{@{}l}{\itshape Correspondence controls (privileged, EEG-only inference)} \\
Cond.-matched shuffle & 0.775\,\ci{0.726}{0.821} & 2.6\,\ci{0.4}{4.9} & 70.5 & 70.0 \\
Unrestricted shuffle & 0.748\,\ci{0.697}{0.796} & 5.3\,\ci{2.2}{8.4}$^{*}$ & 67.9 & 67.5 \\
Random frozen teacher & 0.757\,\ci{0.707}{0.805} & 4.4\,\ci{1.5}{7.4}$^{*}$ & 69.6 & 69.3 \\
\midrule
fMRI teacher (ref.) & 0.836\,\ci{0.793}{0.875} & -- & 75.9 & 75.6 \\
\bottomrule
\end{tabular}
\end{table}

\section{Method}

Band2Net connects a frozen fMRI teacher to the EEG student through a transport
plan estimated individually for each paired observation of a subject
(Fig.~\ref{fig:arch}). Training
involves both modalities; inference uses only the EEG branch.

\subsection{Frequency-specific EEG graphs}
\label{ssec:eeg}

Each preprocessed recording is band-pass filtered into the five canonical bands
$b\in\{\delta,\theta,\alpha,\beta,\gamma\}$. Connectivity in band $b$ for the
$N_E$ channels is computed with the weighted phase-lag index~\cite{vinck2011},
which suppresses the zero-lag coupling added by volume conduction, giving $A^{E}_{b}\in\mathbb{R}^{N_E\times N_E}$. Each channel takes
its row of the unthresholded matrix as node feature, $X^{E}_{b}=A^{E}_{b}$;
spectral power is kept aside as a separate baseline.
Keeping the strongest $25\%$ of off-diagonal entries per observation as
weighted edges, $\tilde A^{E}_{b}$, a density fixed before any test and shared
with the fMRI graphs, produces one graph
$G^{E}_{b}=(\tilde A^{E}_{b},X^{E}_{b})$ per band.

One encoder $E_{E}$ serves all five bands, so features share one space. Band identity is restored via the
learned embedding $p_b$, giving $e_{b}=E_{E}(G^{E}_{b})+p_{b}$ and
$E=[e_{\delta},\dots,e_{\gamma}]\in\mathbb{R}^{5\times d}$. An attention
mechanism scores each band, $s_{b}=\langle q,We_{b}\rangle$ with
$\beta_{b}=\mathrm{softmax}_b(s_{b})$, performs the weighted sum
$z_{E}=\sum_{b}\beta_{b}e_{b}$, and a classifier outputs
$\hat{y}_{S}=g_{S}(z_{E})$.

\subsection{Canonical functional-network teacher}

Regional BOLD signals are extracted with a Schaefer parcellation carrying Yeo
network labels~\cite{schaefer2018,yeo2011}. Functional connectivity between each
pair of the $N_F$ ROIs is the Fisher-transformed Pearson correlation of their
time series, each region takes its whole-brain profile as a node feature, and
edges are thresholded on absolute values as in Sec.~\ref{ssec:eeg}. A compact
ROI encoder $E_{F}$ gives ROI embeddings
$H^{F}\in\mathbb{R}^{N_F\times d}$, where each of the $K$ canonical networks
aggregates the embeddings as
\begin{equation}
  f_{k}=\mathrm{Pool}(\{H^{F}_{i}: i\in\mathcal{N}_{k}\}),
  \;\;
  F=[f_{1},\dots,f_{K}]\in\mathbb{R}^{K\times d},
\end{equation}
where $\mathcal{N}_{k}$ represents the ROIs belonging to network $k$ and
$\mathrm{Pool}$ is the mean. The teacher
classifier $g_{F}$ classifies $\mathrm{Pool}(f_{1},\dots,f_{K})$ under a
cross-entropy loss; each fMRI run yields its own tokens. One teacher is
learned per outer fold and seed ($25$ in all) on the outer-training participants outside the inner
validation set (Sec.~\ref{ssec:protocol}) and then frozen; it supplies the
tokens for all outer-training participants within that fold, so that $f_{k}$
lives in one latent space and is comparable between subjects.

\subsection{Transport distillation}

Only the EEG representation is constrained to be compatible with the teacher.
The projector $P_{E}$, a $64$--$128$--$64$ MLP, maps each band embedding into
the teacher space,
$u_{b}=P_{E}(e_{b})$. With $\bar u_b=u_b/\lVert u_b\rVert$ and
$\bar v_k=f_k/\lVert f_k\rVert$ for the frozen tokens, the cost for paired
observation $s$ (one participant's EEG and fMRI from the same session and
context; eight per participant, $2$ sessions $\times$ $4$ contexts, never pooled
across participants) is $C^{(s)}_{bk}=1-\bar u^{(s)\top}_{b}\bar v^{(s)}_{k}$. Because
the teacher is frozen, the cost cannot be lowered by moving both
representations together without retaining correspondence.

We state this as a balanced entropic transport problem over
$\mathcal{U}(a,b)=\{\Pi\ge 0:\Pi\mathbf{1}_{K}=a,\ \Pi^{\top}\mathbf{1}_{5}=b\}$,
with uniform marginals $a=\tfrac{1}{5}\mathbf{1}_{5}$ and
$b=\tfrac{1}{K}\mathbf{1}_{K}$. The per-observation plan is
\begin{equation}
  \Pi^{*(s)}=\operatorname*{arg\,min}_{\Pi\in\mathcal{U}(a,b)}
  \langle\Pi,C^{(s)}\rangle
  +\varepsilon\sum_{b,k}\Pi_{bk}\!\left(\log\Pi_{bk}-1\right),
\end{equation}
solved by batched log-domain Sinkhorn iterations~\cite{cuturi2013}
without gradient tracking.
With fixed
marginals each band sends and each network receives equal mass, so the plan
only describes \emph{where} supervision goes; a band's influence on the
decision is set by the readout weights $\beta_{b}$.

Rows of $\Pi^{*}$ sum to $\tfrac{1}{5}$, so the transported target for band $b$ is
the row-normalised linear combination $t_{b}=5\sum_{k}\Pi^{*}_{bk}f_{k}$, giving
one privileged fMRI target per band rather than one for the whole recording. The
student is then trained with
\begin{equation}
  \mathcal{L}=\mathcal{L}_{\mathrm{CE}}
  +\frac{\lambda_{\mathrm{OT}}}{5}\sum_{b=1}^{5}
  \left\lVert P_{E}(e_{b})-\mathrm{sg}(t_{b})\right\rVert_{2}^{2}.
  \label{eq:loss}
\end{equation}
The cosine cost decides only where mass is routed; the loss compares
unnormalised vectors, so the student also matches the scale of its target.
The stop-gradient covers the whole target path: $C$, $\Pi^{*}$ and $t_b$
are recomputed from the current embeddings but are constants in the backward
pass, so gradients reach the student only through $P_E(e_b)$ and
$\mathcal{L}_{\mathrm{CE}}$. Plans can shift as the student learns, but each
$t_b$ is a convex combination of fixed teacher tokens and the column constraint
gives every token equal mass, so targets stay within the convex hull of the
tokens and no network can be ignored. At inference the fMRI branch, the
projector and the transport solver are discarded.

\begin{figure*}[t]
  \centering
  \includegraphics[width=0.78\textwidth]{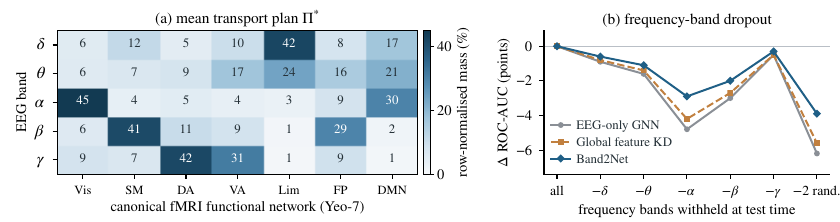}
  \caption{(a) Mean transport plan over training
  observations, folds and seeds, normalised per band (\%, rounded); seed-to-seed
  total variation $0.11$. Yeo-7 networks as in Fig.~\ref{fig:arch}:
  visual, somatomotor, dorsal and ventral attention, limbic, frontoparietal,
  default mode. (b) Change in ROC-AUC points ($\mathrm{AUC}\times100$) when bands
  are withheld at test time, relative to each model's full-band score.}
  \label{fig:results}
\end{figure*}

\section{Experiments}

\subsection{Data and preprocessing}

PsiConnect is a public multimodal study of context-dependent brain
dynamics under psilocybin, released through
OpenNeuro~\cite{novelli2026,stoliker2026,openneuro_ds006110}.
Each participant was recorded before and after a $19$\,mg oral dose of
psilocybin, in four contexts per session: eyes-closed rest, guided
meditation, music listening and movie viewing. EEG and fMRI were acquired
separately under matched conditions, not simultaneously, so supervision is
state-matched rather than time-locked. Of the $65$ participants listed, $56$ gave
usable EEG and fMRI both pre- and post-administration after quality control,
producing $56\times2\times4=448$ paired observations, balanced by construction
across conditions. A session failed if any fMRI run had mean framewise
displacement above $0.5$\,mm, or if RELAX rejected over $20\%$ of channels or
left fewer than $120$ clean $2$-s epochs in any EEG context; a participant was
kept only if both sessions passed in both modalities. Brain--MGF~\cite{yap2025brainmgf}
fuses both modalities on this cohort but evaluates only rest and meditation on
a different sample, so its figures are not comparable.

fMRI was multi-echo EPI ($\mathrm{TR}=0.91$\,s), processed with
fMRIPrep~\cite{esteban2019} using motion correction, normalisation to MNI space,
echo combination and band-pass filtering to $0.01$--$0.1$\,Hz. Signals were then
averaged within the $200$ parcels of the Schaefer atlas, each labelled with one of
the seven Yeo networks~\cite{schaefer2018}, so $N_F=200$ and $K=7$; one run
covers each context. EEG was recorded from $64$ scalp channels at $500$\,Hz,
with separate EOG and ECG channels, and cleaned with RELAX~\cite{bailey2023relax};
the four contexts were identified from the continuous recording by stimulus
markers. After removing the EOG and ECG channels, interpolating rejected scalp
channels ($N_E=64$) and re-referencing to the average, the bands $\delta$ $0.5$--$4$, $\theta$ $4$--$8$,
$\alpha$ $8$--$13$, $\beta$ $13$--$30$ and $\gamma$ $30$--$45$\,Hz were extracted
by zero-phase fourth-order Butterworth filtering, with gamma cut off at
$45$\,Hz to avoid the $50$\,Hz line. wPLI used the Hilbert
cross-spectrum over the first $120$ clean, non-overlapping $2$-s epochs of each
context, equalising data length and estimator bias across contexts.

\subsection{Evaluation protocol}
\label{ssec:protocol}

The subject is the unit of observation. Subject-grouped five-fold
cross-validation over five fixed seeds ensures that no subject appears
in both test and training sets through another session, context or modality, and
batches are balanced across contexts. In each outer fold, $20\%$ of the
training participants form an inner validation set. The teacher and every graph
student are fitted on the rest and early-stopped on inner-validation
cross-entropy; the frozen teacher then supplies tokens for all outer-training
participants, and outer-test fMRI is never used for training or model
selection. For each method separately, $\varepsilon\in\{0.05,0.1,0.2\}$
(transport variants), the distillation weight $\lambda\in\{0.1,0.3,1\}$ (for
Band2Net and every KD baseline; $\lambda_{\mathrm{OT}}$ in Eq.~\ref{eq:loss}),
$\tau\in\{0.1,1,10\}$ (unbalanced plan) and the LR/SVM penalty
$C_{\mathrm{reg}}\in\{0.01,0.1,1,10\}$ are selected in every outer fold and seed
by inner-validation AUC on that fold's training participants only; each
pipeline is fixed before it predicts outer-test EEG. Across the $25$ splits,
Band2Net's modal choices were $\varepsilon=0.1$ ($14$) and
$\lambda_{\mathrm{OT}}=0.3$ ($13$).

Each seed yields one
out-of-fold prediction per session context; these are averaged over seeds and
then over contexts, giving one score for each of the $112$ sessions, so no
subject is duplicated in metrics or resampling. Metrics are computed once on
these averaged probabilities (Platt-scaled for SVMs), never averaged over folds
or seeds; balanced accuracy and macro-F1 use a threshold of $0.5$, fixed in
advance as the classes are balanced. Uncertainty comes from
bootstrapping participants with replacement (both sessions together, never
context-level rows), giving
$95\%$ percentile intervals ($10{,}000$ resamples). Paired
comparisons apply the same resample to both models, never treating folds as
independent observations. Band2Net vs.\ the EEG-only student is the single
primary comparison, fixed in advance; all other paired comparisons are
secondary with unadjusted intervals; Holm correction over the $16$ secondary
comparisons in Table~\ref{tab:main}, applied post hoc as a robustness check, is
marked there, and the four context-shift gains are left unadjusted. The
context-shift test reuses
these folds, so test participants are unseen, and removes the held-out context
from all training, teacher included, and from hyperparameter selection.
As condition is partly confounded with acquisition order and day, we speak of
\emph{session-condition decoding}, not a pharmacological biomarker.

\subsection{Baselines and implementation}

Per-band relative power with logistic regression tests how far simple
spectral features solve the problem, vectorised wPLI connectivity with a
linear SVM whether graph learning helps, and tangent-space features of
shrinkage-regularised band-wise covariances with logistic
regression~\cite{barachant2012} give the standard Riemannian EEG baseline. The
EEG-only GNN uses the proposed encoder and classifier with the cross-entropy
loss alone. Logit~\cite{hinton2015}, global feature~\cite{romero2015fitnets}
and contrastive representation distillation (CRD)~\cite{tian2020crd} to the
pooled teacher embedding represent classical privileged learning, while
pooled-target matching replaces $\Pi^{*}$ with the uniform plan
$\Pi_{bk}=1/(5K)$, sending the same pooled teacher vector to every band.

Six ablations change only the plan or the tokens. The fixed random plan solves
the balanced problem once for a random cost and keeps it throughout training
(four draws), giving band-specific targets without learned routing. The
row-softmax plan, $\Pi_{bk}=\tfrac{1}{5}\mathrm{softmax}_k(-C_{bk}/\varepsilon)$,
drops the column constraint. The shared plan applies one Sinkhorn plan, computed
from the running mean of the cost over training observations, to every
observation, so routing cannot adapt to participant, session or context. The
unbalanced plan relaxes the column constraint with a KL penalty of weight
$\tau$~\cite{chizat2018}. Descriptor targets replace the learned
tokens with label-free descriptors: network $k$ is described by its mean
connectivity to every network, $d_k\in\mathbb{R}^{K}$, mapped to $d=64$ by a
fixed random orthonormal map. Cross-fitted tokens come from teachers fitted
without the participant (four inner folds).

Three variants probe the correspondence. The condition-matched shuffle permutes
teacher tokens among training participants within each condition $\times$
context cell (one derangement per fold and seed), removing individual
correspondence but keeping condition and context; the unrestricted shuffle
permutes across all training observations, so it also mismatches condition for
about half of them. The randomly initialised frozen teacher is never fitted and
carries no learned fMRI content.

All privileged variants share the EEG encoder, classifier and schedule. The encoders are two graph convolutional layers
(symmetric normalisation, self-loops) with
residual connections, layer normalisation and $0.3$ dropout, hidden width $128$
and output dimension $d=64$, with mean node pooling. Training uses AdamW
with learning rate
$3\times10^{-4}$, weight decay $10^{-4}$, batch size $16$, at most $200$
epochs and early stopping after
$20$ epochs without inner-validation improvement; Sinkhorn runs for $50$
iterations. Withheld bands are dropped before the attention
readout, whose
softmax renormalises over the rest. Experiments used one NVIDIA
A100 (40\,GB) GPU. Code will be released upon publication.

\section{Results and discussion}

In Table~\ref{tab:main}, hand-crafted features are informative (band power
$0.694$, wPLI $0.728$, Riemannian $0.739$), while the EEG-only GNN ($0.751$)
is the strongest EEG-only model. Band2Net reaches $0.801$ AUC, $73.2\%$
balanced accuracy and $72.8\%$ macro-F1, with a primary gain of $5.0$ AUC points
\ci{1.8}{8.2} over the EEG-only student. Pooled-target and global feature
distillation perform similarly ($0.772$ vs.\ $0.768$); nominal margins over
these baselines suggest a possible benefit from band-specific routing. Only
margins over band power, wPLI, the unrestricted shuffle and the random teacher
survive Holm correction.

\looseness=-1
Learned routing among distinct teacher tokens (mean pairwise cosine $0.47$) nominally adds
beyond target diversity ($2.7$ AUC points \ci{0.4}{5.1}, not Holm-significant);
the row-softmax margin is also nominal ($2.5$ AUC points \ci{0.3}{4.7}),
and KL relaxation gives $0.791$. The shared-plan difference ($1.4$ AUC points
\ci{$-$0.7}{3.5}) remains unresolved, consistent with similar variation between
and within participants ($F$ ratio $1.16$, permutation $p=$\,$0.23$).
Label-free network descriptors reach $0.793$ (gap $0.8$ AUC points
\ci{$-$1.4}{3.1}), recovering most of the gain through network connectivity
summaries without a learned teacher. Band-specific targets offer little apparent
benefit without learned routing (fixed random plan $0.774$ vs.\ pooled-target
$0.772$), whereas a single learned plan shared across observations performs close
to per-observation routing ($0.787$ vs.\ $0.801$).

The condition-matched shuffle falls $2.6$ AUC points \ci{0.4}{4.9}
below Band2Net, a nominal advantage suggesting a benefit from pairing each
participant's EEG with their own fMRI. The unrestricted shuffle ($0.748$) falls below
the EEG-only student because it also corrupts condition supervision, and a
randomly initialised frozen teacher ($0.757$) shows that the alignment
term and architecture alone explain little of the gain. Band2Net's only extra
trainable module, the projector (about $0.02$M parameters), is discarded
at inference. The teacher reaches $0.95$ AUC on its training participants
and $0.836$ on unseen ones, yet cross-fitted tokens, free of each
participant's own label, give $0.794$, so in-sample tokens do not
explain the gain.

Fig.~\ref{fig:results}(a) shows the mean transport plan. It is diffuse rather
than near one-to-one: alpha, beta, gamma and delta concentrate on different
networks and theta is the least committed band; across participants the cells
have a median IQR of $3.3$ percentage points of row mass, and per-observation row entropy falls
from $1.92$ to $1.49$ nats in training. Entries are teacher-space
weights, not physiological coupling; delta's top weight is on the
susceptibility-prone limbic network.

Fig.~\ref{fig:results}(b) withholds frequency bands at test time. Omitting alpha
hurts every model most, consistent with alpha modulation under
psilocybin~\cite{muthukumaraswamy2013}. Band2Net loses $2.9$ AUC points
against $4.8$ for the EEG-only student, and $3.9$ against
$6.2$ when two random bands are withheld, point estimates indicating an
embedding less tied to one band. Omitting gamma costs every model under one
AUC point, possibly reflecting residual muscle artefact at $30$--$45$\,Hz.
Leaving one context out, Band2Net's paired gain over the EEG-only student is $4.2$
\ci{0.6}{7.9} AUC points for rest, $3.6$ \ci{$-$0.1}{7.3} for meditation,
$2.4$ \ci{$-$1.4}{6.2} for music and $0.8$ \ci{$-$3.2}{4.8} for
movie; only the rest interval excludes zero.

The cohort is small, $56$ participants from one data set. Session condition is
partly confounded with acquisition order and day, although predicted
probability does not drift with context position within pre-dose sessions
(slope $0.004$ per context, \ci{$-$0.012}{0.020}); AUC is
similar with and without prior meditation training ($0.809$ vs.\
$0.792$). As the modalities were not recorded simultaneously, the plan
aligns state-level representations, not neurovascular coupling.

\section{Conclusion}

Band2Net transfers fMRI network structure from separately acquired recordings
into EEG-only decoding through balanced band-to-network transport. It improves
decoding by $5.0$ AUC points over an EEG-only student; secondary comparisons
nominally favour it over logit, global-feature, contrastive and pooled-target
distillation. Participant-specific pairing shows a nominal benefit, while
label-free network descriptors recover most of the gain. EEG-only deployment
supports repeated monitoring outside the scanner. Whether gains extend to
order-balanced cohorts and continuous outcomes such as subjective drug
intensity remains open; a single learned plan may suffice at this sample size.

\section{Compliance with ethical standards}

This is a secondary analysis of de-identified, publicly released data (CC0
waiver); the original study was approved by an institutional human research
ethics committee and all participants gave informed consent.

\section{Acknowledgments}

The authors thank the PsiConnect investigators for releasing the data under
an open licence. This research was funded by Khalifa University of Science and
Technology. The authors have no relevant financial or non-financial interests
to disclose.
ChatGPT (OpenAI) assisted with writing and Claude Opus~5.5 (Anthropic) with
coding and revision; the authors reviewed all AI-assisted output and take full
responsibility for the content of this paper.

\bibliographystyle{IEEEbib}
\bibliography{references}

\end{document}